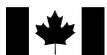

National Research
Council Canada

Conseil national
de recherches Canada

Institute for
Information Technology

Institut de technologie
de l'information

# *Extraction of Keyphrases from Text: Evaluation of Four Algorithms*

P. Turney
October 23, 1997

## Abstract

This report presents an empirical evaluation of four algorithms for automatically extracting keywords and keyphrases from documents. The four algorithms are compared using five different collections of documents. For each document, we have a target set of keyphrases, which were generated by hand. The target keyphrases were generated for human readers; they were not tailored for any of the four keyphrase extraction algorithms. Each of the algorithms was evaluated by the degree to which the algorithm's keyphrases matched the manually generated keyphrases. The four algorithms were (1) the AutoSummarize feature in Microsoft's Word 97, (2) an algorithm based on Eric Brill's part-of-speech tagger, (3) the Summarize feature in Verity's Search 97, and (4) NRC's Extractor algorithm. For all five document collections, NRC's Extractor yields the best match with the manually generated keyphrases.[1]

## Contents

# 1. *Introduction*

This report evaluates four different methods for automatically extracting keywords and keyphrases from documents. Each method is implemented as software that takes an electronic document as input and produces a list of phrases as output. We evaluate the methods by comparing them with keyphrases generated by hand.

By the term *keyphrase*, we mean a list of phrases such as many academic journals include at the beginning of an article. Most journals use the terminology *key word* or *keyword*, but we prefer *keyphrase*, since it seems that about half of all so-called keywords are in fact phrases of two or more words.

The task we consider here is to take a document as input and automatically generate a list (in no particular order) of keyphrases as output. In general, this task could be called *keyphrase generation*, but all four algorithms that we evaluate here perform *keyphrase extraction*. With keyphrase extraction, the output keyphrases always appear somewhere in the body of the input document. Although authors occasionally supply keyphrases that they do not use in the body of their articles, in our corpora, we find that the body of an average document contains 65% (Section 4.6) to 90% (Section 4.3) of the author's keyphrases.

We evaluate the algorithms by comparing their output phrases with target phrases. In the five corpora we investigate here, the target keyphrases are mostly supplied by the authors of the documents. Because the algorithms use keyphrase extraction, in general they cannot achieve 100% agreement with the target phrases, since at least 10% of the target phrases do not appear in the input text. However, generation of good keyphrases that do not appear in the input text is a much more difficult task than extraction of good keyphrases from the input text.

Keyphrase extraction may be viewed as a classification problem. A document can be seen as a bag of phrases, where each phrase belongs in one of two possible classes: either it is a *keyphrase* or it is a *non-keyphrase*. We approach this problem from the perspective of machine learning research. We treat it as a problem of *supervised learning from examples*. We divide our documents into two sets, *training* documents and *testing* documents. The training documents are used to tune the keyphrase extraction algorithms, to attempt to maximize their performance. That is, the training documents are used to teach the supervised learning algorithms how to distinguish keyphrases from non-keyphrases. The testing documents are used to evaluate the tuned algorithms.

The motivation for this work is the range of applications for keyphrases. There are at least five general application areas for keyphrases:

1. **Text Summarization**
   (a) **Mini-summary:** Automatic keyphrase extraction can provide a quick *mini-summary* for a long document. For example, it could be a feature in a web browser; just click the *summarize* button when browsing a long web page, and then a window pops up with the list of keyphrases. It could also be a feature in a word processor. For example, Microsoft's Word 97 has this feature, as we discuss later.
   (b) **Annotated Lists:** Automatic keyphrase extraction can supply added information in a list or table of contents. For example, each item in the *hit list* generated by a web search engine could have a list of keyphrases in addition to the standard information (typically, the URL, title, and first few words). This can be especially helpful for the many web pages that have no title. As another example, a document viewer could use an annotated table of contents to facilitate browsing through the document. The annotations for each section of the document could be generated by running each section through a keyphrase extraction algorithm.

(c) **Labels:** Keyphrases can supply quickly understood labels for documents in a user inter-face where there is a need to display a set of documents (e.g., file manager, mail tool, etc.). It handles the problem that often file names or e-mail subjects are not adequate labels. This can be useful, for example, in applications that use a spatial metaphor, where documents are represented by points in a two- or three-dimensional space. The points could be labelled by their keyphrases.

(d) **Highlights:** It can highlight keyphrases in a long document, to facilitate skimming the document.

(e) **Author Assistance:** Automatic keyphrase extraction can help an author or editor who wants to supply a list of keyphrases for a document. For example, the administrator of a web site might want to have a keyphrase list at the top of each web page. The automati-cally extracted phrases can be a starting point for further manual refinement by the author or editor.

(f) **Text Compression:** On a device with limited display capacity or limited bandwidth, key-phrases can be a substitute for the full text. For example, an email message could be reduced to a set of keyphrases for display on a pager; a web page could be reduced for display on a portable wireless web browser.

2. **Human-Readable Index**

(a) **Journal Index:** Automatic keyphrase extraction can be used to generate a human-read-able index, for access to journal articles or magazine articles. A list of keyphrases, unlike a list of all of the phrases that appear in the articles (a full-text index), will often be small enough for a human reader to scroll through. Thus a keyphrase index can be *browsed*, whereas a full-text index can only be *searched*.

(b) **Resource Access:** It can provide automatic generation of a human-readable index for access to resources, such as a yellow pages, want ads, etc.

(c) **Internet Access:** It can provide automatic generation of a human-readable index for access to web pages, ftp, gopher, etc.

(d) **On-the-fly Indexing:** Automatic keyphrase extraction can generate a human-readable index for a dynamically generated cluster of documents. For example, the hit list of a conventional search engine could be presented as an alphabetical list of keyphrases, instead of a ranked list of titles that match the query.

(e) **Back-of-the-book Index:** It can supply a human-readable index for a book or on-line documentation. Although a list of keyphrases typically contains less than ten items, whereas a back-of-the-book index may contain hundreds or thousands of items, an auto-matic keyphrase extraction algorithm could be used to generate a back-of-the-book index by breaking a long document into a series of short documents. For example, each page or section of a book could be treated as one document.

3. **Interactive Query Refinement**

(a) **Narrow Hit List:** Automatic keyphrase extraction can provide suggestions for improv-ing a query. Often a query with a conventional search engine returns a huge list of match-ing documents. The user would like to narrow the list by adding new terms to the query, but it is not clear what terms should be added. One way to generate suggestions for refin-ing a query is to extract keyphrases from the documents in the hit list for the original query. Conjunction of the new terms with the old query terms yields a shorter hit list. (A new term might be contained in all of the documents in the original hit list, in which case conjunction would not change the hit list, but the list of suggestions could be filtered to

# 1. *Introduction*

remove such terms.)

    (b) **Expand Hit List:** New terms can be added to a query by disjunction, instead of conjunction, which will yield a longer hit list. (Again, filtering may be required.)

4. **Machine-Readable Index**

    (a) **Improved Precision:** It can improve conventional search by improved weighting of terms and phrases in a conventional search engine index table, resulting in more precise retrieval of documents (i.e., fewer irrelevant documents).

    (b) **Compression:** It can be used to compress the index tables that are used by conventional search engines (full-text indexes), which can save memory space in constrained applications (portable computing) or very large applications (indexing very large document collections).

5. **Feature Extraction as Preprocessing for Further Machine Analysis**

    (a) **Document Clustering:** Extracted phrases may be used to generate feature vectors for clustering or hierarchy construction (unsupervised learning of document classes). This is useful for automatic generation of subject hierarchies. (Examples of subject hierarchies are Yahoo, the Library of Congress catalog system, and the Dewey Decimal system.)

    (b) **Document Classification:** Extracted phrases may be used to generate feature vectors for document classification (supervised learning of document classes). This is useful for mail filtering, forwarding documents, and sorting documents.

This list is not intended to be exhaustive, and there may be some overlap in the items. The point of the list is that there are many uses for keyphrases, so a tool for automatically generating good keyphrases should have a sizable market.

    The goal of this report is to provide an objective evaluation of the four keyphrase extraction algorithms. This requires a precise, formal measure of the quality of a list of keyphrases. Section 2 discusses our performance measure in detail. There are (at least) three general criticisms one might make of our method of evaluation. First, it takes human-generated keyphrases as the standard by which machine-generated keyphrases are evaluated. An alternative would be to choose a particular task, such as interactive query refinement, and compare the four algorithms by evaluating them in the context of the given task. We decided against this alternative approach, because there are so many tasks from which to choose. Second, granted that the evaluation is to be based on comparing human-generated and machine-generated keyphrases, there are many ways one might calculate a numerical performance score. In Section 2, we attempt to address this concern by providing a thorough explanation and justification for our scoring method. Third, granted our particular method for calculating a performance score, it could be argued that authors are not qualified to produce good keyphrases. Trained library scientists may be a better source for keyphrases. We examined some documents for which professional indexers had supplied keyphrases, but we found that relatively few of these keyphrases appeared in the bodies of the corresponding documents. It appears that professional indexers tend to generate keyphrases using a substantial amount of information that is external to the given document. Since all four algorithms that we evaluate here are based on keyphrase extraction, rather than unrestricted keyphrase generation, we decided that it would be better to evaluate them by comparison with the authors' keyphrases.

    The four keyphrase extraction algorithms are discussed in Section 3. The four algorithms are (1) the AutoSummarize feature in Microsoft's Word 97, (2) an algorithm based on Eric Brill's part-of-speech tagger, (3) the Summarize feature in Verity's Search 97, and (4) NRC's Extractor algorithm (patent pending).[1] We should point out that only NRC's Extractor is explicitly designed to emulate human-generated keyphrases. Microsoft, Eric Brill, and Verity

might not agree that it is appropriate to evaluate their software in manner of this report. Furthermore, although we are using the AutoSummarize feature in Microsoft Word 97 with no enhancements or alterations, we had to add some post-processing to Eric Brill's part-of-speech tagger and Verity's Search 97, in order to apply them to this task. The results of our experimental evaluations depend critically on this post-processing. Therefore this report is not really an evaluation of Eric Brill's part-of-speech tagger; it is an evaluation of a system that includes Eric Brill's part-of-speech tagger as an essential component. Likewise, this report is not an evaluation of the Summarize feature in Verity's Search 97; it is an evaluation of a system that includes the Summarize feature as an essential component.

The experiments are presented in Section 4. The five corpora are also discussed in this section. Each corpus is described along with its corresponding experimental results. The five document collections are (1) journal articles from five different academic journals, (2) email messages from six NRC employees, (3) web pages from the Aliweb search engine, (4) web pages from NASA, and (5) web pages from FIPS (the US Government's Federal Information Processing Standards). For these five corpora, given our experimental design and performance evaluation metric, NRC's Extractor algorithm has the best performance.

In Section 5, we discuss the interpretation of the experimental results. Section 6 presents related work, and Section 7 is the conclusion.

## 2. *Measuring Performance of the Algorithms*

In this report, we measure the performance of the four keyphrase extraction algorithms by comparing their output to handmade keyphrases. The performance measure is based on the number of matches between the machine-generated phrases and the human-generated phrases. In the following subsections, we define what we mean by *matching* phrases and we describe how the performance measure is calculated from the number of matches.

### 2.1 Criteria for Matching Phrases

If an author suggests the keyphrase "neural network" and a keyphrase generation algorithm suggests the keyphrase "neural networks", we want to count this as a match, although one phrase is singular and the other is plural. On the other hand, if the author suggests "neural networks" and the algorithm suggests "networks", we do not want to count this as a match, since there are many different kinds of networks.

In the experiments that follow, we say that a handmade keyphrase matches a machine-generated keyphrase when they correspond to the same sequence of stems. A stem is what remains when we remove the suffixes from a word.

By this definition, "neural networks" matches "neural network", but it does not match "networks". The order in the sequence is important, so "helicopter skiing" does not match "skiing helicopter". To be more precise about our criteria for matching phrases, we need to say more about how a word is converted to its stem.

### 2.2 Stemming Words

The Porter (1980) and Lovins (1968) stemming algorithms are the two most popular algorithms for stemming English words.[2] Both algorithms use heuristic rules to remove or transform English suffixes. Another approach to stemming is to use a dictionary that explicitly lists the stem for every word that might be encountered in the given text. Heuristics are usually preferred

## 2. *Measuring Performance of the Algorithms*

to a dictionary, due to the labour involved in constructing the dictionary and the computer resources (storage space and execution time) required to use the dictionary.

The Lovins stemmer is more aggressive than the Porter stemmer. That is, the Lovins stemmer is more likely to map two words to the same stem (psychology, psychologist), but it is also more likely to make mistakes (police, policy). We have found that aggressive stemming is better for keyphrase extraction than conservative stemming. In the experiments in this report, we have used an aggressive stemming algorithm that we call the Iterated Lovins stemmer. The algorithm repeatedly applies the Lovins stemmer, until the word stops changing. For example, given "incredible" as input, the Lovins stemmer generates "incred" as output. Given "incred" as input, it generates "incr" as output. With "incr" as input, the output is also "incr". Thus the Iterated Lovins algorithm, given "incredible" as input, generates "incr" as output. Iterating in this manner will necessarily increase (or leave unchanged) the aggressiveness of any stemmer.

Table 1 shows the result of applying the Porter, Lovins, and Iterated Lovins stemmer to some sample words. The Iterated Lovins algorithm recognizes that "jealousness" and "jealousy" share a common stem, whereas the Porter and Lovins stemmers map these two words to distinct stems. On the other hand, the Lovins and Iterated Lovins stemmers map "police" and "policy" to the same stem, whereas the Porter stemmer correctly distinguishes these words.

Table 1: Samples of the behaviour of three different stemming algorithms.

| Word | Porter Stem | Lovins Stem | Iterated Lovins Stem |
|---|---|---|---|
| memory | memori | memor | memor |
| memorable | memor | memor | memor |
| memorize | memor | memor | memor |
| believes | believ | belief | belief |
| belief | belief | belief | belief |
| believable | believ | belief | belief |
| science | scienc | sci | sc |
| scientist | scientist | sci | sc |
| scientific | scientif | scientif | scientif |
| jealousness | jealous | jeal | jeal |
| jealousy | jealousi | jealous | jeal |
| realistic | realist | real | real |
| reality | realiti | re | re |
| incredible | incred | incred | incr |
| incredulous | incredul | incredl | incredl |
| beautiful | beauti | beaut | beaut |
| beauty | beauti | beaut | beaut |
| psychology | psychologi | psycholog | psycholog |
| psychologist | psychologist | psycholog | psycholog |
| police | polic | polic | pol |
| policy | polici | polic | pol |
| assemblies | assembli | assembl | assembl |
| assembly | assembli | assemb | assemb |

### 2.3  Method for Scoring Matches

We may view keyphrase extraction as a classification problem. If we think of a document as a set of words and phrases, then the task is to classify each word or phrase into one of two categories: either it is a keyphrase or it is not a keyphrase. We can evaluate an automatic keyphrase extraction algorithm by the degree to which its classifications correspond to the human-generated classifications. The outcome of applying a keyphrase extraction algorithm to a corpus can be neatly summarized with a *confusion matrix*, as in Table 2. The variable *a* represents the number of times that the human-generated phrase matches a machine-generated phrase. The variable *d* represents the number of times that the human and the machine agree that a phrase is not a keyphrase. The variables *b* and *c* represent the number of times that the human and the machine disagree on the classification of a phrase.

Table 2: The confusion matrix for keyphrase classification.

|  | Classified as a Keyphrase by the Human | Classified as Not a Keyphrase by the Human |
|---|---|---|
| Classified as a Keyphrase by the Machine | *a* | *b* |
| Classified as Not a Keyphrase by the Machine | *c* | *d* |

We consider both "neural network" and "neural networks" to be matches for the phrase "neural network". Therefore it is better to think of the task as classification of *stemmed phrases*, rather than classification of *whole phrases*. The Iterated Lovins stemmer transforms both whole phrases "neural network" and "neural networks" into the stemmed phrase "neur network". If a person suggests that "neural networks" is a good keyphrase for an article, we interpret that as classifying the stemmed phrase "neur network" as a keyphrase.

For examples of the relative values of the variables *a, b, c*, and *d*, consider an average journal article. The average journal article has about 10,000 words (see Table 8). Depending on how we go from words to phrases, this may correspond to about 3,500 unique whole phrases, which in turn may correspond to about 2,500 unique stemmed phrases. Thus $a + b + c + d \approx 2500$. The author supplies an average of 7.5 keyphrases, of which about 82% appear in the full text, so about 6 ($7.5 \cdot 0.82 = 6.15$) of the author's keyphrases match phrases in the text. People never suggest keyphrases that map to the same stemmed phrase. For example, no author would suggest both "neural network" and "neural networks" as keyphrases for the same article. Therefore we may assume that the author's 6 whole phrases map to 6 unique stemmed phrases. Thus $a + c \approx 6$. For every positive example of the target class (*keyphrase*), we have about 400 negative examples (counter-examples; *non-keyphrase*):

$$\frac{a+c}{b+d} \approx \frac{6}{2500-6} \approx \frac{1}{400} \tag{1}$$

The skewed distribution of the two classes makes this a difficult classification problem.

We would like to have a single number that represents the performance of a keyphrase extraction algorithm. In other words, we would like a suitable function that maps *a, b, c,* and *d* to a single value. It is common to use *accuracy* to reduce a confusion matrix to a single value:

## 2. *Measuring Performance of the Algorithms*

$$\text{accuracy} = \frac{a+d}{a+b+c+d} \tag{2}$$

Unfortunately, there are some problems with using accuracy here. One problem is that, because the class distribution is highly skewed, we can achieve very high accuracy by always guessing the majority class. That is, if a trivial keyphrase extraction algorithm always generates an empty set of keyphrases ($a = b = 0$), for any input document, its accuracy would be about 99.76%:

$$\frac{a+d}{a+b+c+d} = \frac{d}{c+d} \approx \frac{2500-6}{2500} = 0.9976 \tag{3}$$

Clearly we do not want to assign such a high performance rating to such a poor algorithm.

Another problem with accuracy is that the space of candidate phrases (of size $a + b + c + d$) is not well defined. It depends on how we convert a document into a set of phrases. Each of the four methods for keyphrase extraction has a different approach to generating candidate phrases; each has a different size space of candidate phrases. The variable $d$ is particularly sensitive to this problem. We can estimate the other three variables by examining the machine-generated keyphrases and the human-generated keyphrases. If we assume that all of the human-generated keyphrases are in the space of candidate phrases, then $a + c$ is simply the number of human-generated keyphrases. If we assume that machine-generated keyphrases never collapse to the same stemmed phrase (e.g., the machine never gives both "neural network" and "neural networks" as output keyphrases for a given input document), then $a + b$ is simply the number of machine-generated keyphrases. Since $a$ is the number of matches (where we define matching as above), we can easily estimate *a, b,* and *c.* However, there is no way to estimate $d$ by examining only the machine-generated keyphrases and the human-generated keyphrases. We may be able to estimate $d$ by examining the given input document, but this estimate will be very sensitive to the method for converting a document into a set of phrases. Therefore, we would like a performance measure that is based only on *a, b,* and *c.*

This issue is familiar to researchers in the field of information retrieval. In the standard paradigm of information retrieval, a user enters a query in a search engine, which searches through a document collection and returns a list of documents, chosen from the collection. It is assumed that some of the documents in the collection are relevant to the query and the rest are irrelevant. The performance of the search engine is evaluated by comparing the output of the search engine to the relevance judgements of a human expert. We may view this as a classification problem, where the two classes are *relevant* and *irrelevant*. This view leads to the confusion matrix in Table 3.

Table 3: The confusion matrix for information retrieval.

|  | Classified as Relevant by the Human | Classified as Irrelevant by the Human |
|---|---|---|
| Classified as Relevant by the Machine | *a* | *b* |
| Classified as Irrelevant by the Machine | *c* | *d* |

When we view information retrieval as a classification problem, we encounter the same difficulties as when we view keyphrase extraction as a classification problem. The number of irrel-evant documents $b + d$ is typically much larger than the number of relevant documents $a + c$, so we have a highly skewed class distribution. Also, it can be difficult to estimate $d$, whereas estimating $a$, $b$, and $c$ is relatively easy. For example, when we use a search engine on the web, we can directly estimate $a$, $b$, and $c$, but $d$ depends on the size of the document collection that is indexed by the search engine, which may be unknown to us, and may fluctuate daily.

Researchers in information retrieval use *precision* and *recall* to evaluate the performance of a search engine:

$$\text{precision} = \frac{a}{a + b} \tag{4}$$

$$\text{recall} = \frac{a}{a + c} \tag{5}$$

Precision is an estimate of the probability that, if a given search engine classifies a document as relevant to a user's query, then it really is relevant. Recall is an estimate of the probability that, if a document is relevant to a user's query, then a given search engine will classify it as relevant.

There is a well-known trade-off between precision and recall. We can optimize one at the expense of the other. For example, if we guess that the entire document collection is relevant, then our recall is necessarily 100%, but our precision will be close to 0%. On the other hand, if we take the document that we are most confident is relevant and guess that only this single document is relevant to the user's query, then our precision might be 100%, but our recall would be close to 0%. We want a performance measure that yields a high score only when precision and recall are balanced. A measure that is widely used in information retrieval is the F-measure (van Rijsbergen, 1979; Lewis, 1995):

$$\text{F-measure} = \frac{2 \cdot \text{precision} \cdot \text{recall}}{\text{precision} + \text{recall}} = \frac{2a}{2a + b + c} \tag{6}$$

The F-measure is never greater than the average of precision and recall:

$$\frac{2 \cdot \text{precision} \cdot \text{recall}}{\text{precision} + \text{recall}} \leq \frac{\text{precision} + \text{recall}}{2} \tag{7}$$

To see why this is true, we begin with two definitions:

$$\text{avg} = \frac{\text{precision} + \text{recall}}{2} \tag{8}$$

$$\Delta = |\text{precision} - \text{avg}| = |\text{recall} - \text{avg}| \tag{9}$$

We can now define the F-measure in terms of the average:

$$\text{F-measure} = \frac{2 \cdot (\text{avg} + \Delta) \cdot (\text{avg} - \Delta)}{(\text{avg} + \Delta) + (\text{avg} - \Delta)} \tag{10}$$

$$= \frac{2(\text{avg}^2 - \Delta^2)}{2\text{avg}} = \text{avg} - \frac{\Delta^2}{\text{avg}} \tag{11}$$

It follows that the F-measure is less than the average by a quantity that is proportional to the square of the imbalance between precision and recall, as represented by $\Delta$. When precision and recall are not balanced, the F-measure is strictly less than the average of precision and recall. When precision and recall are equal, the F-measure equals the average.

# 3. *Four Algorithms for Extracting Keyphrases*

In the following experiments, we evaluate the performance of the keyphrase extraction algorithms using the F-measure. We estimate $a + c$ by the number of human-generated keyphrases, which assumes that all of the human-generated keyphrases are in the space of candidate phrases, and that humans never suggest keyphrases that map to the same stemmed phrase. We estimate $a + b$ by the number of machine-generated keyphrases, which assumes that machines never suggest keyphrases that map to the same stemmed phrase. When calculating the number of matches $a$, we allow at most one machine-generated phrase to match a human-generated phrase. Thus, if the machine does suggest two keyphrases that map to the same stemmed phrase (contrary to our preceding assumption), then at most one of these two keyphrases can count as a match.

# 3. *Four Algorithms for Extracting Keyphrases*

This section discusses the four keyphrase extraction algorithms. We do not discuss the core routines of these algorithms, since they are either proprietary (in the case of Microsoft, Verity, and NRC) or they are discussed elsewhere (in the case of Eric Brill's work). What we describe is how these algorithms were applied to the task of keyphrase extraction, in the experiments in Section 4.

## 3.1 Microsoft's Word 97: The AutoSummarize Feature

Microsoft's Word 97 is a complete word processing software package. In this report, we are only concerned with the AutoSummarize feature in Word 97. The AutoSummarize feature is available from the *Tools* menu. The main function of this feature is to identify important sentences in the document that is being edited. The identified sentences can be highlighted or separated from the remainder of the text. The user can specify a target percentage of the text for AutoSummarize to mark as important.

As a side-effect, when AutoSummarize is used, it also fills in the *Keywords* field of the document's *Properties*. The *Properties* form is available from the *File* menu. AutoSummarize always generates exactly five keyphrases, if the document contains at least five distinct words. The keyphrases are always single words, never phrases of two or more words. They are always in lower case, even when they are abbreviations or proper nouns. There is no way for the user of Word 97 to adjust the AutoSummarize feature. For example, it is not possible to ask it for six keyphrases instead of five.

In the following experiments, we used the five keyphrases produced by AutoSummarize with no further processing. Unlike the other three algorithms, there was no tuning of the AutoSummarize feature to the training data.

In the remainder of this paper, we will use the phrase *Microsoft's Word 97* to mean automatic keyphrase extraction using the AutoSummarize feature in Word 97. By *Microsoft's Word 97*, we do not mean the whole Word 97 software package. Microsoft does not claim that the AutoSummarize feature emulates human-generated keyphrases. Microsoft might not agree that the experimental method that we use in this report is a fair test of the AutoSummarize feature.

## 3.2 Eric Brill's Part-of-Speech Tagger: Frequent Noun Phrases

In our corpora, almost all of the target keyphrases are noun phrases. The following simple pattern matches most of the keyphrases:

$$(NN|NNS|NNP|NNPS|JJ)* (NN|NNS|NNP|NNPS|VBG) \qquad (12)$$

This pattern means zero or more nouns or adjectives $(NN|NNS|NNP|NNPS|JJ)*$ , followed by one final noun or gerund $(NN|NNS|NNP|NNPS|VBG)$ . More precisely, NN represents a singular noun, NNS represents a plural noun, NNP represents a singular proper noun, NNPS represents a plural proper noun, JJ represents an adjective, and VBG represents a gerund.[3] Table 4 shows an example of tagged keyphrases for an article from the journal *Psycoloquy* (one of the articles in the Journal Article corpus, Section 4.3). In this example, only the keyphrase "Bayes' theorem" does not match the pattern (12), due to the possessive POS.

Table 4: An example of tagged keyphrases, taken from the Journal Article corpus.

| Author's Keyphrases | Tagged Keyphrases |
| --- | --- |
| base rate fallacy | base/NN rate/NN fallacy/NN |
| Bayes' theorem | Bayes/NNP '/POS theorem/NN |
| decision making | decision/NN making/VBG |
| ecological validity | ecological/JJ validity/NN |
| ethics | ethics/NNS |
| fallacy | fallacy/NN |
| judgment | judgment/NN |
| probability | probability/NN |

Keyphrases often correspond to frequent noun phrases in the text. This suggests the following algorithm for automatic keyphrase extraction:

1. Tag each word in the given document by its part-of-speech.
2. Look for all sequences of tagged words that match the pattern (12).
3. Count the frequency of each matching sequence.
4. Output the $N$ most frequent matching sequences.

This is essentially the algorithm we use, with some minor enhancements, which we will discuss below.

The most difficult part of this algorithm is part-of-speech tagging. For this purpose, we use Eric Brill's part-of-speech tagger (Brill, 1992, 1993, 1994).[4] We quickly found that there were two problems with the algorithm sketched above. First, the $N$ most frequent matching sequences tend to be almost entirely phrases consisting of a single word. If we want to extract multi-word keyphrases, then the algorithm needs some adjustment. Second, the algorithm does not recognize phrases that are minor variants of each other, such as "market", "markets", "marketing". The following algorithm addresses these two concerns:

1. Use Eric Brill's part-of-speech tagger to tag each word in the given document.
2. Look for all single tagged words that match the pattern (12).
3. Look for all sequences of two tagged words that match the pattern.
4. Look for all sequences of three tagged words that match the pattern.
5. Stem the matching sequences using the Iterated Lovins stemmer (Section 2.2).
6. Count the frequency of each matching stemmed sequence.
7. For each of the *N1* most frequent matching stemmed single words, output the most frequent corresponding single whole word. That is, for a given stemmed single word, output the most frequent un-stemmed word that the Iterated Lovins stemmer transforms into the given stemmed word.

8. For each of the *N2* most frequent matching stemmed sequences of two words, output the most frequent corresponding sequence of two whole words.

9. For each of the *N3* most frequent matching stemmed sequences of three words, output the most frequent corresponding sequence of three whole words.

We selected training documents from the corpora, to tune the values of *N1*, *N2*, and *N3* to maximize the F-measure. Thus this algorithm is a form of supervised learning from examples. To find the best values of *N1, N2,* and *N3* for the training data, we used exhaustive search. We did not consider phrases of four or more words, since they are relatively rare in our corpora. The training and testing are described in more detail in Section 4.2.

In the following, we will refer to this algorithm as *Eric Brill's Tagger*, because the part-of-speech tagger is the most important component of the algorithm. However, the tagger is only one component in a larger system, so the reader should not make inferences from the performance of the whole system to the performance of the tagger. Eric Brill has not suggested that this is an appropriate usage of his tagger.

### 3.3   Verity's Search 97: The Summarize Feature

Verity's Search 97 is a complete text retrieval system, including a search engine, an index builder, and a web crawler. In this report, we are only concerned with the Summarize feature in Search 97. The Summarize feature enables the search engine to display summaries of each document in the hit list, along with the usual information, such as the document title. A summary in Search 97 consists of a list of sentences, with highlighted keyphrases embedded in the sentences. The user can control the number of sentences in a summary, by either specifying the number of sentences desired or the percentage of the source document desired.

To evaluate the keyphrases generated by Search 97, we need to remove them from their surrounding sentences. We used the following procedure:

1. Use Verity's Search 97 to generate a summary with *N* sentences, using the given document as input.

2. Look for all highlighted phrases in the resulting summary.

3. Stem the highlighted phrases using the Iterated Lovins stemmer (Section 2.2).

4. For each unique stemmed phrase, output the first corresponding highlighted phrase in the summary.

The Iterated Lovins stemmer is used to eliminate phrases that are minor variants of other phrases that have already been output. The value of *N* was tuned using training documents, to maximize the F-measure. We used exhaustive search to find the best value of *N* for the training data. Thus this algorithm uses a simple form of supervised learning from examples.

In the following, we refer to this algorithm as *Verity's Search 97*, although Search 97 is only a component of the system. Furthermore, only the Summarize feature of Search 97 is used in the algorithm. Therefore the reader should not make inferences from the performance of this algorithm to the performance of the whole Search 97 software package, or even the performance of the Summarize feature, since we are not using this feature as it was intended to be used.

### 3.4   NRC's Extractor

NRC's Extractor (patent pending) takes a document as input and generates a list of keyphrases as output.[5] The algorithm uses supervised learning from examples. Extractor was trained using the same documents as we used with Eric Brill's Tagger and Verity's Search 97. Extractor is intended to emulate human-generated keyphrases. On most hardware platforms, Extractor can process a typical document in about one second.

## 4. *Experiments*

This section presents the results of our experimental evaluation of the four algorithms. We begin with a description of the five document collections that were used in the experiments and the experimental design. We then discuss each document collection separately, along with the experimental results for the collection. We end with a summary of the results.

### 4.1 Document Collections

The experiments in this report are based on five different document collections, listed in Table 5. For each document, there is a target set of keyphrases, generated by hand. In the Journal Article corpus, the keyphrases were created by the authors of the articles. In the Email corpus, a university student created the keyphrases. In the three web page corpora, the keyphrases were created by the authors of the web pages (as far as we know).

Table 5: The five document collections.

| Corpus Name | Description | Size |
|---|---|---|
| Journal Articles | articles from five different academic journals | 75 |
| Email Messages | email messages from six different NRC employees | 311 |
| Aliweb Web Pages | web pages from the Aliweb search engine | 90 |
| NASA Web Pages | web pages from NASA's Langley Research Center | 141 |
| FIPS Web Pages | web pages from the US Government's Federal Information Processing Standards | 35 |

### 4.2 Design of Experiments

As we discussed in Section 3, three of the four algorithms require training. We used 55 articles from the Journal Article corpus and 235 messages from the Email Message corpus as training data (roughly 75% of each collection). The remaining journal articles and email messages, and all of the web pages, were used for testing.

The training of NRC's Extractor involves a random component. That is, each time Extractor is trained, it learns a slightly different model of the training data. Therefore we ran Extractor 10 times on each training collection. In the following results, we present the average and the standard deviation of the 10 runs. There is no random component to the training of Eric Brill's Tagger or Verity's Search 97, so these algorithms were only run once on each training collection. There is no training at all for Microsoft's Word 97.

The three trainable algorithms were trained separately for the Journal Article corpus and the Email Message Corpus. These two corpora have quite different characteristics (as we discuss below), so we did not merge them. The most significant difference between them is the length of the individual documents. The average journal article has 10,781 words, whereas the average email message has 376 words. When testing with the 20 test articles from the Journal Article corpus, the three trainable algorithms used the models that they induced from the 55 training articles. When testing with the 76 test messages from the Email Message corpus, the three trainable

## 4. *Experiments*

algorithms used the models that they induced from the 235 training messages. When testing with the three web page corpora, we alternated between the Journal Article model and the Email model, based on the length of the given web page. We used a threshold of 20K bytes to choose the appropriate model. This threshold was chosen because almost all of the email messages are shorter than 20K and almost all of the journal articles are longer than 20K. When a given web page was shorter than 20K, the trainable algorithm used the model that it had induced from the Email corpus. When the given web page was longer than 20K, the trainable algorithm used the model that it had induced from the Journal Article corpus. This procedure was applied uniformly to all three trainable algorithms (Eric Brill's Tagger, Verity's Search 97, and NRC's Extractor).

In all cases, the author's keyphrases were removed from the documents before the text was presented to the keyphrase extraction algorithms. The author's keyphrases were used only for tuning performance on the training documents and evaluating performance on the testing documents.

### 4.3   Journal Articles

We selected 75 journal articles from five different journals, listed in Table 6. The full text of each article is available on the web. The authors have supplied keyphrases for each of these articles.

Table 6: Sources for the journal articles.

| Documents | Journal Name and URL |
|---|---|
| 1-6 | *Journal of the International Academy of Hospitality Research*<br>http://borg.lib.vt.edu/ejournals/JIAHR/jiahr.html |
| 7-26 | *Psycoloquy*<br>http://www.princeton.edu/~harnad/psyc.html |
| 27-28 | *The Neuroscientist*<br>http://www.theneuroscientist.com/ |
| 29-42 | *Journal of Computer-Aided Molecular Design*<br>http://www.ibc.wustl.edu/jcamd/ |
| 43-75 | *Behavioral & Brain Sciences Preprint Archive*<br>http://www.princeton.edu/~harnad/bbs.html |

Three of the journals are about cognition (*Psycoloquy, The Neuroscientist, Behavioral & Brain Sciences Preprint Archive*), one is about the hotel industry (*Journal of the International Academy of Hospitality Research*), and one is about chemistry (*Journal of Computer-Aided Molecular Design*). This mix of journals lets us see whether there is significant variation in performance of the keyphrase extraction algorithms among journals in the same field (cognition) and among journals in different fields (cognition, hotel industry, chemistry). Our experiments indicate that the chemistry journal is particularly challenging for automatic keyphrase extraction.

In Table 7, we show the distribution of words per keyphrase for the 75 journal articles. Most authors use one or two words in a keyphrase. Occasionally they will use three words, but only rarely will authors use four or five words. Table 8 gives some indication of how the statistics vary among the five journals. Table 9 shows, for each journal, the percentage of the authors' keyphrases that appear at least once in the full text of the article (excluding the keyword list, of course).

Table 7: Number of words per keyphrase for the Journal Article corpus.

| Words per Keyphrase | Percent of Keyphrases |
|---|---|
| 1 | 46.5% |
| 2 | 44.5% |
| 3 | 7.5% |
| 4 | 1.1% |
| 5 | 0.4% |
| 1 to 5 | 100.0% |

Table 8: Some statistics for each of the five journals.

| Journal Name | Number of Articles | Average Number of … ± Standard Deviation | | |
|---|---|---|---|---|
| | | Keyphrases per Document | Words per Keyphrase | Words per Document |
| *Journal of the International Academy of Hospitality Research* | 6 | 6.2 ± 2.6 | 2.0 ± 0.8 | 6,299 ± 2,066 |
| *Psycoloquy* | 20 | 8.4 ± 3.1 | 1.5 ± 0.6 | 4,350 ± 2,726 |
| *The Neuroscientist* | 2 | 6.0 ± 1.4 | 1.8 ± 1.1 | 7,476 ± 856 |
| *Journal of Computer-Aided Molecular Design* | 14 | 4.7 ± 1.4 | 1.9 ± 0.6 | 6,474 ± 2,633 |
| *Behavioral & Brain Sciences Preprint Archive* | 33 | 8.4 ± 2.2 | 1.6 ± 0.7 | 17,522 ± 6,911 |
| All Five Journals | 75 | 7.5 ± 2.8 | 1.6 ± 0.7 | 10,781 ± 7,807 |

Table 9: Percentage of the authors' keyphrases that appear in the corresponding full text.

| Journal Name | Keyphrases in Full Text |
|---|---|
| *Journal of the International Academy of Hospitality Research* | 70.3% |
| *Psycoloquy* | 74.9% |
| *The Neuroscientist* | 91.7% |
| *Journal of Computer-Aided Molecular Design* | 78.8% |
| *Behavioral & Brain Sciences Preprint Archive* | 87.4% |
| All Five Journals | 81.6% |

In the following experiments, *Psycoloquy* was used as the testing set and the remaining journals were used as the training set. We chose this split because it resulted in roughly 75% training cases and 25% testing cases. We did not use a random split, because we wanted to test the ability of the learning algorithms to generalize across journals. A random 75/25 split would most likely have resulted in a training set with samples of articles from all five journals. We wanted the testing set to contain articles from a journal that was not represented in the training set.

Table 10 shows the training and testing performance of the four algorithms. Note that the Word 97 does not require training, but the other three algorithms were tuned with the training documents, as described in Section 3. Table 11 breaks out the training set performance for the four different journals. It is clear that the *Journal of Computer-Aided Molecular Design* is much

## 4. Experiments

more challenging than the other journals. This is likely due to the very specialized vocabulary of chemistry.

Table 10: Performance of the four algorithms on the Journal Article corpus.

| Keyphrase Extraction Algorithm | Training Set Performance 55 Articles | | | Testing Set Performance 20 Articles | | |
|---|---|---|---|---|---|---|
| | Precision | Recall | F-measure | Precision | Recall | F-measure |
| Microsoft's Word 97 | 0.098 | 0.069 | 0.081 | 0.170 | 0.102 | 0.127 |
| Eric Brill's Tagger | 0.149 | 0.189 | 0.167 | 0.167 | 0.180 | 0.173 |
| Verity's Search 97 | 0.119 | 0.163 | 0.138 | 0.168 | 0.180 | 0.173 |
| NRC's Extractor | 0.208 ± 0.020 | 0.228 ± 0.015 | 0.216 ± 0.008 | 0.226 ± 0.019 | 0.224 ± 0.026 | 0.223 ± 0.014 |

Table 11: Performance of the four algorithms on each of the five journals.

| Keyphrase Extraction Algorithm | F-measure on Training Set Journals | | | | Testing |
|---|---|---|---|---|---|
| | Hospitality 6 Articles | Neuroscientist 2 Articles | Molecular 14 Articles | Brain 33 Articles | Psycoloquy 20 Articles |
| Microsoft's Word 97 | 0.060 | 0.091 | 0.015 | 0.104 | 0.127 |
| Eric Brill's Tagger | 0.176 | 0.133 | 0.073 | 0.199 | 0.173 |
| Verity's Search 97 | 0.182 | 0.229 | 0.080 | 0.145 | 0.173 |
| NRC's Extractor | 0.238 ± 0.032 | 0.374 ± 0.059 | 0.133 ± 0.015 | 0.233 ± 0.009 | 0.223 ± 0.014 |

The experimental results show that NRC's Extractor achieves the best performance on this corpus, about 30% better than the nearest competitors ($0.223/0.173 = 1.289$). Eric Brill's Tagger and Verity's Search 97 are tied on the testing set articles. Microsoft's Word 97 has the lowest score for this corpus.

Table 12 shows the output keyphrases for the four algorithms and the authors' keyphrases, for the first three articles from the journal *Psycoloquy*. In this table, where the machine's output matches the author's output, the matching phrase is italicized. The training of Extractor has a random component, which is why the performance of Extractor is averaged over 10 separate training runs. The examples of the output of Extractor, in Table 12 below, are chosen from the run of the 10 training runs that is most typical (most like the average of the 10 runs; not the best run). In this most typical run, the F-measure for *Psycoloquy* was 0.224, compared to an average of 0.223 (see Table 10).

Note that, for the second article, "Brain Rhythms, Cell Assemblies and Cognition: Evidence from the Processing of Words and Pseudowords," the author's keyphrase "cell assembly" does not match Extractor's keyphrase "cell assemblies", because the Iterated Lovins stemmer maps "assembly" to "assemb", but "assemblies" maps to "assembl" (see Table 1). These kinds of problems are inevitable with any purely mechanical performance measure. However, we believe that the benefits of mechanical performance measures are greater than the costs. Mechanical performance measures lack human knowledge of semantics, but they are precise, objective, repeatable, unbiased, fast, and simple.

Table 12: Examples of the output of the algorithms for *Psycoloquy*.

| Source of Keyphrases | Keyphrases | F-measure |
|---|---|---|
| 1. | "The Base Rate Fallacy Myth" | |
| Author's Keyphrases: | base rate fallacy, Bayes' theorem, decision making, ecological validity, ethics, fallacy, judgment, probability. | |
| Microsoft's Word 97: | rate, base, information, *judgment*, psychology. | 0.154 |
| Eric Brill's Tagger: | rates, base, base rates, information, Psychology, *judgments*, decision, Social Psychology, decision makers. | 0.118 |
| Verity's Search 97: | *base rate fallacy, judgment*, base rates, subjects, *probability*. | 0.462 |
| NRC's Extractor: | base rates, information, psychology, *judgments*, *probability*, *base rate fallacy*, experiments. | 0.400 |
| 2. | "Brain Rhythms, Cell Assemblies and Cognition: Evidence from the Processing of Words and Pseudowords" | |
| Author's Keyphrases: | brain theory, cell assembly, cognition, event related potentials, ERP, electroencephalograph, EEG, gamma band, Hebb, language, lexical processing, magnetoencephalography, MEG, psychophysiology, periodicity, power spectral analysis, synchrony. | |
| Microsoft's Word 97: | assembly, neuron, word, activity, process. | 0.000 |
| Eric Brill's Tagger: | neurons, activity, words, assembly, responses, processing, *cell assembly*, cell assemblies, spectral power. | 0.077 |
| Verity's Search 97: | cortical cell assemblies, word processing, Cell assembly activity, responses, activity, neurons, gamma-band responses, words, pseudowords. | 0.000 |
| NRC's Extractor: | cell assemblies, neurons, activity, words, cognitive processing, gamma-band responses, *EEG*, Pulvermueller. | 0.080 |
| 3. | "On the Evolution of Consciousness and Language" | |
| Author's Keyphrases: | consciousness, language, plans, motivation, evolution, motor system. | |
| Microsoft's Word 97: | *plan*, *consciousness*, process, *language*, action. | 0.545 |
| Eric Brill's Tagger: | *plans*, action, *consciousness*, process, psychology, *language*, planning mechanism, New York, episodic memory. | 0.400 |
| Verity's Search 97: | Psychology, *plans*, organize, *Consciousness*, plan-executing, behavior, actions, *Language*. | 0.429 |
| NRC's Extractor: | *plans*, *consciousness*, *language*, behavior, planning mechanism, organization, communication. | 0.462 |

## 4.4   Email Messages

We collected 311 email messages from six different NRC employees. Most of the messages were incoming mail, both internal NRC mail and external mail. We believe that these messages are representative of typical messages that are exchanged in corporate and institutional environments. A university student created the keyphrases for the 311 email messages.[6] The student was asked to create the kind of keyphrases that are used for journal articles. We tried to avoid biasing the student's keyphrases in favour of any particular method for automatic keyphrase extraction, but we did encourage the student to use phrases that appear in the body or subject field of the

## 4. *Experiments*

corresponding email message. Since one individual created all of the keyphrases, this corpus is likely to be more homogenous than the journal article corpus or the web page corpus.

Table 13 shows the distribution of words per keyphrase for the email corpus. Like the journal article corpus, most keyphrases have one to three words. The keyphrases in the email corpus tend to be slightly longer than the keyphrases in the journal corpus. It seems likely that this is a reflection of the tastes of the student, rather than an intrinsic property of the email corpus. Table 14 indicates that there is relatively little variation in the statistical properties of the messages among the six employees. Table 15 shows that the student complied closely to the request that the keyphrases should appear in the message.

Table 13: Number of words per keyphrase for the Email Message corpus.

| Words per Keyphrase | Percent of Keyphrases with Given Number of Words | | |
|---|---|---|---|
| | Training Set 235 Messages | Testing Set 76 Messages | Whole Corpus 311 Messages |
| 1 | 47.5% | 44.3% | 46.7% |
| 2 | 34.0% | 34.8% | 34.2% |
| 3 | 12.3% | 12.8% | 12.4% |
| 4 | 4.8% | 7.1% | 5.3% |
| 5 to 8 | 1.4% | 1.0% | 1.4% |
| 1 to 8 | 100.0% | 100.0% | 100.0% |

Table 14: Some statistics for each of the six employees.

| Employee | Number of Messages | Average Number of … ± Standard Deviation | | |
|---|---|---|---|---|
| | | Keyphrases per Document | Words per Keyphrase | Words per Document |
| #1 | 47 | $6.6 \pm 5.0$ | $2.0 \pm 1.1$ | $542 \pm 606$ |
| #2 | 41 | $7.7 \pm 6.7$ | $1.7 \pm 0.8$ | $328 \pm 374$ |
| #3 | 42 | $4.3 \pm 3.8$ | $1.5 \pm 0.8$ | $454 \pm 698$ |
| #4 | 96 | $3.6 \pm 2.1$ | $1.8 \pm 1.0$ | $230 \pm 243$ |
| #5 | 41 | $4.6 \pm 4.3$ | $1.6 \pm 0.8$ | $453 \pm 805$ |
| #6 | 44 | $3.9 \pm 2.5$ | $2.1 \pm 1.1$ | $413 \pm 674$ |
| All Six Employees | 311 | $4.9 \pm 4.3$ | $1.8 \pm 1.0$ | $376 \pm 561$ |

The data were randomly split into testing and training sets, by randomly selecting, for each employee, 75% of the messages for training and 25% for testing. Table 13 shows that the training data have essentially the same statistical characteristics as the testing data.

Table 16 shows the training and testing set performance of the four algorithms on the email corpus. Again, NRC's Extractor is about 30% ahead of the nearest competitor ($0.225 / 0.174 = 1.293$). Eric Brill's Tagger is close to Verity's Search 97, and Microsoft's Word 97 has the lowest performance.

Table 15: Percentage of the keyphrases that appear in the corresponding email message.

| Employee | Number of Messages | Keyphrases in Email |
|----------|-------------------|---------------------|
| #1 | 47 | 97.4% |
| #2 | 41 | 97.5% |
| #3 | 42 | 98.9% |
| #4 | 96 | 97.1% |
| #5 | 41 | 98.9% |
| #6 | 44 | 98.8% |
| All Six Employees | 311 | 97.9% |

Table 16: Performance of the four algorithms on the Email Message corpus.

| Keyphrase Extraction Algorithm | Training Set Performance 235 Messages | | | Testing Set Performance 76 Messages | | |
|---|---|---|---|---|---|---|
| | Precision | Recall | F-measure | Precision | Recall | F-measure |
| Microsoft's Word 97 | 0.109 | 0.111 | 0.110 | 0.145 | 0.149 | 0.147 |
| Eric Brill's Tagger | 0.147 | 0.205 | 0.171 | 0.141 | 0.196 | 0.164 |
| Verity's Search 97 | 0.143 | 0.214 | 0.171 | 0.145 | 0.217 | 0.174 |
| NRC's Extractor | 0.209 ± 0.016 | 0.268 ± 0.016 | 0.234 ± 0.008 | 0.200 ± 0.022 | 0.258 ± 0.018 | 0.225 ± 0.018 |

## 4.5 Aliweb Web Pages

We collected 90 web pages using the Aliweb search engine, a public search engine provided by NEXOR Ltd. in the UK.[7] Most web search engines use a *spider* to collect web pages for their index. A spider is a program that gathers web pages by expanding an initial list of URLs by following the hypertext links on the corresponding web pages. Aliweb is unusual in that it does not use a spider to collect web pages; instead, it has an electronic fill-in form, where people are asked to enter any URLs that they would like to add to the Aliweb index. Among other things, this fill-in form has a field for keyphrases. The keyphrases are stored in the Aliweb index, along with the URLs.

We did a *substring* search with Aliweb, using the query string "e" (the most common letter in English text) and setting the maximum number of matches at 1000. The intent of this search was to gather a relatively large, random sample of web pages. We looked at each entry in the list of search results and deleted duplicate entries, entries with no corresponding keyphrases, and entries with clearly poor keyphrases. We were left with 90 distinct web pages with associated keyphrases. It is our impression that the web pages were typically submitted to Aliweb by their authors, so most of the keyphrases were probably supplied by the authors of the web pages.

Table 17 shows the distribution of words per keyphrase for the web page corpus. No keyphrases contain more than three words. In this corpus, keyphrases tend to contain fewer words than in either the journal article corpus or the email corpus. Table 18 shows some statistics for the corpus. Note that the web page corpus does not have an internal structure, in the sense that the journal article corpus separates into five journals and the email corpus separates into six recipients. About 70% of the keyphrases appear in the full text of the corresponding web pages.

# 4. *Experiments*

Table 17: Number of words per keyphrase for the Aliweb Web Page corpus.

| Words per Keyphrase | Percent of Keyphrases |
|:---:|:---:|
| 1 | 81.0% |
| 2 | 16.6% |
| 3 | 2.4% |
| 1 to 3 | 100.0% |

Table 18: Some statistics for the Aliweb Web Page corpus.

| Description of Statistic | Value of Statistic |
|:---|:---:|
| Average Number of Keyphrases per Document ± Standard Deviation | $6.0 \pm 3.0$ |
| Average Number of Words per Keyphrase ± Standard Deviation | $1.2 \pm 0.5$ |
| Average Number of Words per Document ± Standard Deviation | $949 \pm 2603$ |
| Percentage of the Keyphrases that Appear in the Full Text | 69.0% |

In the following experiments, this corpus is used for testing only. Thus there is no division of the data into testing and training sets. The keyphrases in this corpus seem subjectively to be of lower quality than the keyphrases in the other corpora. Although they are useful for testing purposes, they do not seem suitable for training.

Table 19 shows the performance of the four algorithms on the web page corpus. Again NRC's Extractor has the best performance, but it is only 3% ahead of Microsoft's Word 97 ($0.227 / 0.220 = 1.032$). The target keyphrases for this corpus are mainly (81% — see Table 17) single words, which is ideal for Word 97, since it can only generate single word phrases. Verity's Search 97 has the third highest score, followed by Eric Brill's Tagger.

Table 19: Performance of the four algorithms on the Aliweb Web Page corpus.

| Keyphrase Extraction Algorithm | Corpus: 90 Web Pages | | |
|:---|:---:|:---:|:---:|
| | Precision | Recall | F-measure |
| Microsoft's Word 97 | 0.241 | 0.201 | 0.220 |
| Eric Brill's Tagger | 0.128 | 0.153 | 0.140 |
| Verity's Search 97 | 0.192 | 0.200 | 0.196 |
| NRC's Extractor | 0.237 | 0.218 | 0.227 |

## 4.6  NASA Web Pages

We collected 141 web pages from NASA's Langley Research Center. Their Technology Applications Group (TAG) has 141 web pages that describe technology they have developed.[8] The web pages are intended to attract the interest of potential industrial partners and customers. Each page includes a list of keyphrases.

Table 20 shows the number of words per keyphrase. This corpus has relatively more two- and three-word keyphrases than the other corpora. Table 21 shows that the documents are relatively short and relatively fewer keyphrases can be found in the bodies of the corresponding documents. This may explain why the four algorithms perform less well on this collection.

Table 20: Number of words per keyphrase for the NASA Web Page corpus.

| Words per Keyphrase | Percent of Keyphrases |
|:---:|:---:|
| 1 | 34.1% |
| 2 | 45.0% |
| 3 | 16.7% |
| 4 to 7 | 4.2% |
| 1 to 7 | 100.0% |

Table 21: Some statistics for the NASA Web Page corpus.

| Description of Statistic | Value of Statistic |
|:---|:---:|
| Average Number of Keyphrases per Document ± Standard Deviation | $4.7 \pm 2.0$ |
| Average Number of Words per Keyphrase ± Standard Deviation | $1.9 \pm 0.9$ |
| Average Number of Words per Document ± Standard Deviation | $466 \pm 102$ |
| Percentage of the Keyphrases that Appear in the Full Text | 65.3% |

We used this corpus for testing only. Table 22 shows the performance of the four algorithms. NRC's Extractor has the highest score, followed very closely by Eric Brill's Tagger. Microsoft's Word 97 performed poorly because there were relatively few one-word phrases (34% — see Table 20).

Table 22: Performance of the four algorithms on the NASA Web Page corpus.

| Keyphrase Extraction Algorithm | Corpus: 141 Web Pages | | |
|:---|:---:|:---:|:---:|
| | Precision | Recall | F-measure |
| Microsoft's Word 97 | 0.084 | 0.089 | 0.086 |
| Eric Brill's Tagger | 0.120 | 0.179 | 0.143 |
| Verity's Search 97 | 0.109 | 0.177 | 0.135 |
| NRC's Extractor | 0.135 | 0.156 | 0.145 |

## 4.7   FIPS Web Pages

We have gathered 35 web pages from the US government's Federal Information Processing Standards (FIPS).[9] These documents define the standards to which US government departments must conform when purchasing computer hardware and software. Each document includes a list of keyphrases.

Table 23 shows the distribution of words per keyphrase in this corpus. There is an unusual number of four-word keyphrases, because almost every document includes the keyphrase "Federal Information Processing Standard". If we ignore this phrase, the distribution is similar to the distribution in the Email Message corpus. From Table 24, we can see that the documents are relatively long and that many of the keyphrases appear in the body of the corresponding document.

This corpus was used for testing only. Table 25 shows the performance of the four algorithms. NRC's Extractor scored 6% above the nearest competitor, Verity's Search 97 $(0.208 / 0.196 = 1.061)$.

## 5. *Discussion*

Table 23: Number of words per keyphrase for the FIPS Web Page corpus.

| Words per Keyphrase | Percent of Keyphrases |
|:---:|:---:|
| 1 | 43.7% |
| 2 | 31.9% |
| 3 | 9.2% |
| 4 | 13.0% |
| 5 | 2.2% |
| 1 to 5 | 100.0% |

Table 24: Some statistics for the FIPS Web Page corpus.

| Description of Statistic | Value of Statistic |
|:---|:---:|
| Average Number of Keyphrases per Document ± Standard Deviation | $9.0 \pm 3.5$ |
| Average Number of Words per Keyphrase ± Standard Deviation | $2.0 \pm 1.1$ |
| Average Number of Words per Document ± Standard Deviation | $7025 \pm 6385$ |
| Percentage of the Keyphrases that Appear in the Full Text | 78.2% |

Table 25: Performance of the four algorithms on the FIPS Web Page corpus.

| Keyphrase Extraction Algorithm | Corpus: 35 Web Pages | | |
|:---|:---:|:---:|:---:|
| | Precision | Recall | F-measure |
| Microsoft's Word 97 | 0.246 | 0.136 | 0.175 |
| Eric Brill's Tagger | 0.167 | 0.152 | 0.159 |
| Verity's Search 97 | 0.193 | 0.199 | 0.196 |
| NRC's Extractor | 0.241 | 0.184 | 0.208 |

### 4.8 Summary of Results

Table 26 summarizes the performance of the four algorithms on the five corpora. Only the testing set results appear in the table. On average, NRC's Extractor is 18% ahead of the nearest competitor, Verity's Search 97 ($0.206 / 0.175 = 1.177$). NRC's Extractor has the highest score on all five corpora, although the margin is very slim on the NASA Web Page corpus. Verity's Search 97 is in second place on three corpora, tying with Eric Brill's Tagger on one of these three.

## 5. *Discussion*

In this section, we give our interpretation of the experimental results. One way to view the results in Table 26 is to consider the performance of each algorithm relative to NRC's Extractor, as we do in Table 27. In this table, we have shaded in grey the second highest performance for each corpus. We see that Extractor is ahead by a large margin on the Journal Article and Email Message corpora. These two corpora are most like the training documents (Section 4.2). On the remaining three corpora, the margin between Extractor and second place is much smaller. These three corpora are relatively unlike the training documents.

The results show that Extractor performs particularly well when the testing documents are like the training documents. However, even when the testing documents are unlike the training documents, Extractor performs at least as well as the best competing algorithm. Thus it seems that Extractor is not overfitting the training data.

Table 26: Summary of performance of the four algorithms on the five corpora.

| Test Corpus | Corpus Size | F-measure | | | |
|---|---|---|---|---|---|
| | | Microsoft's Word 97 | Eric Brill's Tagger | Verity's Search 97 | NRC's Extractor |
| Journal Articles (Testing Set Only) | 20 | 0.127 | 0.173 | 0.173 | 0.223 |
| Email Messages (Testing Set Only) | 76 | 0.147 | 0.164 | 0.174 | 0.225 |
| Aliweb Web Pages | 90 | 0.220 | 0.140 | 0.196 | 0.227 |
| NASA Web Pages | 141 | 0.086 | 0.143 | 0.135 | 0.145 |
| FIPS Web Pages | 35 | 0.175 | 0.159 | 0.196 | 0.208 |
| Average | 72 | 0.151 | 0.156 | 0.175 | 0.206 |

Table 27: Performance of the algorithms relative to Extractor.

| Test Corpus | F-measure as Percentage of NRC's Extractor | | | |
|---|---|---|---|---|
| | Microsoft's Word 97 | Eric Brill's Tagger | Verity's Search 97 | NRC's Extractor |
| Journal Articles (Testing Set Only) | 57% | 78% | 78% | 100% |
| Email Messages (Testing Set Only) | 65% | 73% | 77% | 100% |
| Aliweb Web Pages | 97% | 62% | 86% | 100% |
| NASA Web Pages | 59% | 99% | 93% | 100% |
| FIPS Web Pages | 84% | 76% | 94% | 100% |
| Average | 73% | 77% | 86% | 100% |

The results suggest that a wider range of training documents may enable Extractor to increase its lead over the competing algorithms. The other two learning algorithms (Eric Brill's Tagger and Verity's Search 97) are also likely to benefit from an increased range of training documents. In our experiments, the three learning algorithms each induced two models, one for the Journal Article corpus and one for the Email Message corpus. During testing, the size of the input document determined the selection of the model (Section 4.2). Short documents were processed with the Email Message model and long documents were processed with the Journal Article model. If we train the three learning algorithms with a wider range of documents, then we will need a more sophisticated method for selecting the appropriate model. Ideally a single model would be able to handle all varieties of documents. This is an area for future work.

On the surface, it might seem that our analysis is based on a relatively small sample. We have a total of 362 testing documents (summing across the five corpora). However, the appropriate level of granularity here is phrases, not documents. Viewed as a supervised learning problem, the task is to classify each phrase as either *keyphrase* or *non-keyphrase*. In Table 28 we show the size of the testing collection according to three different measures. The second column (corpus size) lists the number of documents in each corpus. The third column (number of words) lists the number of words in each corpus. This total includes stop words (e.g., "the", "if") and counts each repetition of a word. The fourth column is an estimate of the number of unique phrases in the documents. It is only an estimate, because the number depends on what we mean by a *phrase*. Each of the four keyphrase extraction algorithms contains a different definition of phrase, implicit in the algorithm. We do not include repetitions of a phrase within a single document, but we do include repetitions of a phrase across multiple documents. We estimate the number of unique phrases as 25% of the number of (non-unique) words. Table 28 shows that our analysis is actually based on a relatively large sample size, by the current standards of the machine learning research community.

Table 28: Various ways of looking at the size of the testing data.

| Test Corpus | Corpus Size | Total Number of Words (Including Duplicates) | Total Number of Phrases (Excluding Duplicates) |
|---|---|---|---|
| Journal Articles (Testing Set Only) | 20 | 87,000 | 21,750 |
| Email Messages (Testing Set Only) | 76 | 28,600 | 7,150 |
| Aliweb Web Pages | 90 | 85,400 | 21,350 |
| NASA Web Pages | 141 | 65,700 | 16,425 |
| FIPS Web Pages | 35 | 246,000 | 61,500 |
| Total | 362 | 512,700 | 128,175 |

# 6. *Related Work*

In this section, we discuss some related work. Although there are several papers that discuss automatically extracting important phrases, none of these papers treat this problem as supervised learning from examples.

Krulwich and Burkey (1996) use heuristics to extract significant phrases from a document. The heuristics are based on syntactic clues, such as the use of italics, the presence of phrases in section headers, and the use of acronyms. Their motivation is to produce phrases for use as features when automatically classifying documents. Their algorithm tends to produce a relatively large list of phrases, so it has low precision, and thus low F-measure.

Muñoz (1996) uses an unsupervised learning algorithm to discover two-word keyphrases. The algorithm is based on Adaptive Resonance Theory (ART) neural networks. Muñoz's algorithm tends to produce a large list of phrases, so it has low precision, and thus low F-measure. Also, the algorithm is not applicable to one-word or more-than-two-word keyphrases.

Steier and Belew (1993) use the mutual information statistic to discover two-word keyphrases. This approach has the same limitations as Muñoz (1996), when considered as a keyphrase extraction algorithm: it produces a low precision list of two-word phrases. Steier and Belew (1993) compare the mutual information of word pairs within specific topic areas (e.g., documents concerned with labour relations) and across more general collections (e.g., legal documents). They make the interesting observation that certain phrases that would seem to be highly characteristic of a certain topic area (e.g., "union member" would seem to be characteristic of documents concerned with labour relations) actually have a higher mutual information statistic across more general collections (e.g., "union member" has a higher mutual information across a general legal collection than within the topic area of labour relations).

Several papers explore the task of producing a summary of a document by extracting key sentences from the document (Luhn, 1958; Edmundson, 1969; Marsh *et al.,* 1984; Paice, 1990; Paice and Jones, 1993; Johnson *et al.,* 1993; Salton *et al.,* 1994; Kupiec *et al.,* 1995; Brandow *et al.,* 1995; Jang and Myaeng, 1997). This task is similar to the task of keyphrase extraction, but it is more difficult. The extracted sentences often lack cohesion because anaphoric references are not resolved (Johnson *et al.,* 1993; Brandow *et al.,* 1995). *Anaphors* are pronouns (e.g., "it", "they"), definite noun phrases (e.g., "the car"), and demonstratives (e.g., "this", "these") that refer to previously discussed concepts. When a sentence is extracted out of the context of its neighbouring sentences, it may be impossible or very difficult for the reader of the summary to determine the referents of the anaphors. Johnson *et al.* (1993) attempt to automatically resolve anaphors, but their system tends to produce overly long summaries. Keyphrase extraction avoids this problem because anaphors are not keyphrases.[10] Also, a list of keyphrases has no structure; unlike a list of sentences, a list of keyphrases can be randomly permuted without significant consequences.[11]

Most of these papers on summarization by sentence extraction describe algorithms that are based on manually derived heuristics. The heuristics tend to be effective for the intended domain, but they often do not generalize well to a new domain. Extending the heuristics to a new domain involves a significant amount of manual work. A few of the papers describe learning algorithms, which can be trained by supplying documents with associated target summaries (Kupiec *et al.,* 1995; Jang and Myaeng, 1997). Learning algorithms can be extended to new domains with less work than algorithms that use manually derived heuristics. However, there is still some manual work involved, because the training summaries must be composed of sentences that appear in the document, which means that standard author-supplied abstracts are not suitable. An advantage of keyphrase extraction is that standard author-supplied keyphrases are suitable for training a learning algorithm, because the majority of such keyphrases appear in the bodies of the corresponding documents. Kupiec *et al.* (1995) and Jang and Myaeng (1997) use a Bayesian statistical model to learn how to extract key sentences. A Bayesian approach may be applicable to keyphrase extraction.

Another body of related work addresses the task of *information extraction.* An information extraction system seeks specific information in a document, according to predefined guidelines. The guidelines are specific to a given topic area. For example, if the topic area is news reports of terrorist attacks, the guidelines might specify that the information extraction system should identify (i) the terrorist organization involved in the attack, (ii) the victims of the attack, (iii) the type of attack (kidnapping, murder, etc.), and other information of this type that can be expected in a typical document in the topic area. ARPA has sponsored a series of *Message Understanding Conferences* (MUC-3, 1991; MUC-4, 1992; MUC-5, 1993; MUC-6, 1995), where information extraction systems are evaluated with corpora in various topic areas, including terrorist attacks and corporate mergers.

Most information extraction systems are manually built for a single topic area, which requires a large amount of expert labour. The highest performance at the Fifth Message Understanding Conference (MUC-5, 1993) was achieved at the cost of two years of intense programming effort. However, recent work has demonstrated that a learning algorithm can perform as well as a manually constructed system (Soderland and Lehnert, 1994). Soderland and Lehnert (1994) use decision tree induction as the learning component in their information extraction system. We may view the predefined guidelines for a given topic area as defining a template to be

## 6. *Related Work*

filled in by the information extraction system. In Soderland and Lehnert's (1994) system, each slot in the template is handled by a group of decision trees that have been trained specially for that slot. The nodes in the decision trees are based on syntactical features of the text, such as the presence of certain words.

Information extraction and keyphrase extraction are at opposite ends of a continuum that ranges from detailed, specific, and domain-dependent (information extraction) to condensed, general, and domain-independent (keyphrase extraction). The different ends of this continuum require substantially different algorithms. However, there are intermediate points on this continuum. An example is the task of identifying corporate names in business news. This task was introduced in the Sixth Message Understanding Conference (MUC-6, 1995), where it was called the *Named Entity Recognition* task. The competitors in this task were evaluated using the F-measure. The best system achieved a score of 0.964, which indicates that named entity recognition is easier than keyphrase extraction (Krupka, 1995). This system used hand-crafted linguistic rules to recognize named entities.[12]

Other related work addresses the problem of automatically creating an index (Fagan, 1987; Salton, 1988; Ginsberg, 1993; Nakagawa, 1997; Leung and Kan, 1997). Leung and Kan (1997) provide a good survey of this work. There are two general classes of indexes: indexes that are intended for human readers to browse (often called *back-of-book* indexes) and indexes that are intended for use with information retrieval software (*search engine* indexes). Search engine indexes are not suitable for human browsing, since they usually index every occurrence of every word (excluding stop words, such as "the" and "of") in the document collection. Back-of-book indexes tend to be much smaller, since they index only important occurrences of interesting words and phrases.

Search engine indexes often contain single words, but not multi-word phrases. Several researchers have experimented with extending search engine indexes with multi-word phrases. The result of these experiments is that multi-word phrases have little impact on the performance of search engines (Fagan, 1987; Croft, 1991). They do not appear to be worth the extra effort required to generate them.

Since we are interested in keyphrases for human browsing, back-of-book indexes are more relevant than search engine indexes. Leung and Kan (1997) address the problem of learning to assign index terms from a controlled vocabulary. This involves building a statistical model for each index term in the controlled vocabulary. The statistical model attempts to capture the syntactic properties that distinguish documents for which the given index term is appropriate from documents for which it is inappropriate. Their results are interesting, but the use of a controlled vocabulary makes it difficult to compare their work with the algorithms we examine here. We studied a small sample of controlled index terms in the INSPEC database, and we found that very few of these terms appear in the bodies of the corresponding documents.[13] It seems that algorithms that are suitable for automatically generating controlled index terms are substantially different from algorithms that are suitable for automatically extracting keyphrases. It is also worth noting that a list of controlled index terms must grow every year, as the body of literature grows, so Leung and Kan's (1997) software would need to be continuously trained.

Nakagawa (1997) automatically extracts simple and compound nouns from technical manuals, to create back-of-book indexes. Each compound noun is scored using a formula that is based on the frequency of its component nouns in the given document. In his experiments, Nakagawa (1997) evaluates his algorithm by comparing human-generated indexes to machine-generated indexes. He uses van Rijsbergen's (1979) E-measure, which is simply 1 minus the F-measure that we use in our experiments. His E-measure, averaged over five different manuals, corre-sponds to an F-measure of 0.670. This suggests that back-of-book indexes are easier to generate than keyphrases. Two factors that complicate the comparison are that Nakagawa (1997) uses Japanese text, whereas we use English text, and Nakagawa's (1997) human-generated indexes were generated with the assistance of his algorithm, which tends to bias the results in favour of his algorithm.

The main feature that distinguishes a back-of-book index from a keyphrase list is length. As Nakagawa (1997) observes, a document is typically assigned $10^0 \sim 10^1$ keyphrases, but a back-of-book index typically contains $10^2 \sim 10^3$ index terms. Also, keyphrases are usually intended to cover the whole document, but index terms are intended to cover only a small part of a document. A keyphrase extraction algorithm might be used to generate a back-of-book index by breaking a long document into sections of one to three pages each. A back-of-book index generation algorithm might be used to generate keyphrases by selecting index terms that appear on many pages throughout the book. Another distinguishing feature is that a sophisticated back-of-book index is not simply an alphabetical list of terms. There is often a hierarchical structure, where a major index term is followed by an indented list of related minor index terms.

## 7. Conclusion

We argued in the Introduction that there are many applications for keyphrases. However, a quick estimate suggests that less than half of all journals ask their authors to supply keyphrases. On the web, keyphrase are very rare. Some web pages use the META tag in HTML to provide keyphrases, but these phrases are typically intended to be read only by automated web crawlers, collecting pages for search engines. In general, the META tag is used to provide a long list of weakly relevant terms, designed to increase the likelihood that the given page will appear near the top of the hit list for any weakly related query. These are not *keyphrases* in the sense that we mean here; they are not suitable for many of the applications we considered in the Introduction.

Since most documents do not currently have associated keyphrases, and it is generally not economical to add keyphrases by hand, there is a need for automatic keyphrase extraction algorithms. We believe that NRC's Extractor, while not perfect, is performing at a level that makes it suitable for many of the applications that we listed. This report provides evidence that, if you need automatic keyphrase extraction, Extractor is the best choice.

## Notes

1. Microsoft and Word 97 are trademarks or registered trademarks of Microsoft Corporation. Verity and Search 97 are trademarks or registered trademarks of Verity Inc.
2. We used an implementation of the Porter (1980) stemming algorithm written in Perl, by Jim Richardson, at the University of Sydney, Australia. This implementation includes some extensions to Porter's original algorithm, to handle British spelling. It is available at http://www.maths.usyd.edu.au:8000/jimr.html. For the Lovins (1968) stemming algorithm, we used an implementation written in C, by Linh Huynh. This implementation is part of the MG (Managing Gigabytes) search engine, which was developed by a group of people in Australia and New Zealand. The MG code is available at http://www.kbs.citri.edu.au/mg/.
3. These part-of-speech tagging conventions were developed for the Penn Treebank Project. This project is the work of the Linguistic Data Consortium, which is based at the University of Pennsylvania. More information is available at http://www.ldc.upenn.edu/ldc/.
4. Eric Brill's part-of-speech tagger is written in C. It is available at http://www.cs.cmu.edu/afs/

## 7. *Conclusion*

       cs/project/ai-repository/ai/areas/nlp/parsing/taggers/brill/0.html    and   also   at   ftp:// ftp.cs.jhu.edu/pub/brill/Programs/.

5.  A demonstration version of NRC's Extractor (patent pending) is available at http:// ai.iit.nrc.ca/II_public/extractor/. The demonstration version has been trained already; it does not allow the user to make any adjustments. Extractor has been licensed to Tetranet Software Inc. For more information about Tetranet Software, see http://www.tetranetsoftware.com/.

6.  Thanks to Elaine Sin of the University of Calgary for generating the keyphrases and for her help with the experiments with the email corpus.

7.  The Aliweb search engine is available at http://www.nexor.com/public/aliweb/search/doc/ form.html.

8.  The NASA Langley web pages are available at http://tag-www.larc.nasa.gov/tops/ tops_text.html.

9.  The FIPS web pages are available at http://www.itl.nist.gov/div897/pubs/.

10. There may be some exceptions, such as the use of the phrase "the Mob" to refer to an international crime organization.

11. Some journals ask their authors to order their keyphrases from most general to most specific. In this report, we have ignored the order of the keyphrases. For most of the applications we have considered here (Section 1), the order is not important. NRC's Extractor attempts to order the keyphrases it produces from most important to least important.

12. It is now available as a commercial product, called NetOwl Extractor, from IsoQuest. See http://www.isoquest.com/.

13. The INSPEC database is the leading English bibliographic database for scientific and technical literature in physics, electrical engineering, electronics, communications, control engineering, computers and computing, and information technology. It is produced by the Institution of Electrical Engineers. Records in the INSPEC database have fields for both controlled vocabulary index terms (called *descriptors*) and free index terms (called *identifiers*). More information is available at http://www.iee.org.uk/publish/inspec/inspec.html.

# 7. *Conclusion*